\documentclass[11pt,a4paper]{article}
\usepackage[hyperref]{acl2019}
\usepackage{acl2019}
\usepackage{times}
\usepackage{latexsym}
\usepackage{url}
\usepackage{multirow}
\usepackage{adjustbox}
\usepackage[ruled]{algorithm2e} 
\usepackage{amssymb}
\usepackage{pifont}

\usepackage{url}

\usepackage{mwe}
\usepackage{amsmath}
\usepackage{amssymb}
\usepackage{wasysym}
\usepackage{bbm}
\usepackage{todonotes} 
\usepackage{graphicx}
\usepackage{caption}
\usepackage[caption=false]{subfig}

\aclfinalcopy 
\def\aclpaperid{1963} 

\DeclareGraphicsExtensions{.pdf,.png,.jpg}
\newcommand\MNAME{MT-DNN}

\title{Multi-Task Deep Neural Networks for Natural Language Understanding}

\author{Xiaodong Liu\thanks{~~Equal Contribution.}~$^1$, Pengcheng He$^\bold{\ast}$$^2$, Weizhu Chen$^2$, Jianfeng Gao$^1$ \\
  $^1$ Microsoft Research~~~~~~~~
  $^2$ Microsoft Dynamics 365 AI \\
  {\tt \{xiaodl,penhe,wzchen,jfgao\}@microsoft.com}
}
\date{}

\begin{document}
\maketitle

\begin{abstract}
In this paper, we present a Multi-Task Deep Neural Network (MT-DNN) for learning representations across multiple natural language understanding (NLU) tasks. MT-DNN not only leverages large amounts of cross-task data, but also benefits from a regularization effect that leads to more general representations to help adapt to new tasks and domains. MT-DNN extends the model proposed in \citet{liu2015mtl} by incorporating a pre-trained bidirectional transformer language model, known as BERT \citep{bert2018}. 
MT-DNN obtains new state-of-the-art results on ten NLU tasks, including SNLI, SciTail, and eight out of nine GLUE tasks, pushing the GLUE benchmark to 82.7\% (2.2\% absolute improvement) \footnote{As of February 25, 2019 on the latest GLUE test set.}. 
We also demonstrate using the SNLI and SciTail datasets that the representations learned by MT-DNN allow domain adaptation with substantially fewer in-domain labels than the pre-trained BERT representations.
The code and pre-trained models are publicly available at https://github.com/namisan/mt-dnn.
\end{abstract}

\section{Introduction}
\label{sec:introduction}

Learning vector-space representations of text, e.g., words and sentences, is fundamental to many natural language understanding (NLU) tasks. Two popular approaches are \emph{multi-task learning} and \emph{language model pre-training}. In this paper we combine the strengths of both approaches by proposing a new Multi-Task Deep Neural Network (MT-DNN).

Multi-Task Learning (MTL) is inspired by human learning activities where people often apply the knowledge learned from previous tasks to help learn a new task \citep{caruana1997multitask,zhang2017survey}. For example, it is easier for a person who knows how to ski to learn skating than the one who does not. Similarly, it is useful for multiple (related) tasks to be learned jointly so that the knowledge learned in one task can benefit other tasks.  
Recently, there is a growing interest in applying MTL to representation learning using deep neural networks (DNNs) \citep{collobert2011natural,liu2015mtl,luong2015multi,mt-mrc2018,guo2018soft, ruder122019latent} for two reasons. 
First, supervised learning of DNNs requires large amounts of task-specific labeled data, which is not always available. MTL provides an effective way of leveraging supervised data from many related tasks. 
Second, the use of multi-task learning profits from a regularization effect via alleviating overfitting to a specific task, thus making the learned representations universal across tasks. 

In contrast to MTL, language model pre-training has shown to be effective for learning universal language representations by leveraging large amounts of unlabeled data. A recent survey is included in \citet{gao2018neural}. Some of the most prominent examples are ELMo \citep{elmo2018}, GPT \citep{gpt2018} and BERT \citep{bert2018}. These are neural network language models trained on text data using unsupervised objectives. 
For example, BERT is based on a multi-layer bidirectional Transformer, and is trained on plain text for masked word prediction and next sentence prediction tasks. 
To apply a pre-trained model to specific NLU tasks, we often need to fine-tune, for each task, the model with additional task-specific layers using task-specific training data. 
For example, \citet{bert2018} shows that BERT can be fine-tuned this way to create state-of-the-art models for a range of NLU tasks, such as question answering and natural language inference.

We argue that MTL and language model pre-training are complementary technologies, and can be combined to improve the learning of text representations to boost the performance of various NLU tasks.
To this end, we extend the MT-DNN model originally proposed in \citet{liu2015mtl} by incorporating BERT as its shared text encoding layers. 
As shown in Figure 1, the lower layers (i.e., text encoding layers) are shared across all tasks, while the top layers are task-specific, combining different types of NLU tasks such as single-sentence classification, pairwise text classification, text similarity, and relevance ranking. 
Similar to the BERT model, MT-DNN can be adapted to a specific task via fine-tuning. Unlike BERT, MT-DNN uses MTL, in addition to language model pre-training, for learning text representations.

MT-DNN obtains new state-of-the-art results on eight out of nine NLU tasks
\footnote{The only GLUE task where MT-DNN does not create a new state of the art result is WNLI. But as noted in the GLUE webpage (https://gluebenchmark.com/faq), there are issues in the dataset, and none of the submitted systems has ever outperformed the majority voting baseline whose accuracy is 65.1.}
used in the General Language Understanding Evaluation (GLUE) benchmark \citep{wang2018glue}, pushing the GLUE benchmark score to 82.7\%, amounting to 2.2\% absolute improvement over BERT. We further extend the superiority of MT-DNN to the SNLI \cite{bowman2015large} and SciTail \cite{scitail} tasks. The representations learned by MT-DNN allow domain adaptation with substantially fewer in-domain labels than the pre-trained BERT representations. For example, our adapted models achieve the accuracy of 91.6\% on SNLI and 95.0\% on SciTail, outperforming the previous state-of-the-art performance by 1.5\% and 6.7\%, respectively. Even with only 0.1\% or 1.0\% of the original training data, the performance of MT-DNN on both SNLI and SciTail datasets is better than many existing models. All of these clearly demonstrate MT-DNN's exceptional generalization capability via multi-task learning.


\section{Tasks}
\label{sec:tasks}

The MT-DNN model combines four types of NLU tasks: single-sentence classification, pairwise text classification, text similarity scoring, and relevance ranking. For concreteness, we describe them using the NLU tasks defined in the GLUE benchmark as examples. 

\paragraph{Single-Sentence Classification:} 
Given a sentence\footnote{In this study, a sentence can be an arbitrary span of contiguous text or word sequence, rather than a linguistically plausible sentence.}, the model labels it using one of the pre-defined class labels. For example, the \textbf{CoLA} task  is to predict whether an English sentence is grammatically plausible. The \textbf{SST-2} task is to determine whether the sentiment of a sentence extracted from movie reviews is positive or negative.

\paragraph{Text Similarity:}
This is a regression task. Given a pair of sentences, the model predicts a real-value score indicating the semantic similarity of the two sentences. \textbf{STS-B} is the only example of the task in GLUE. 

\paragraph{Pairwise Text Classification:} 
Given a pair of sentences, the model determines the relationship of the two sentences based on a set of pre-defined labels. 
For example, both \textbf{RTE} and \textbf{MNLI} are language inference tasks, where the goal is to predict whether a sentence is an \emph{entailment}, \emph{contradiction}, or \emph{neutral} with respect to the other. 
\textbf{QQP} and \textbf{MRPC} are paraphrase datasets that consist of sentence pairs. The task is to predict whether the sentences in the pair are semantically equivalent.

\paragraph{Relevance Ranking:}
Given a query and a list of candidate answers, the model ranks all the candidates in the order of relevance to the query. 
\textbf{QNLI} is a version of Stanford Question Answering Dataset \citep{rajpurkar2016squad}. 
The task involves assessing whether a sentence contains the correct answer to a given query. 
Although QNLI is defined as a binary classification task in GLUE, in this study we formulate it as a pairwise ranking task, where the model is expected to rank the candidate that contains the correct answer higher than the candidate that does not. 
We will show that this formulation leads to a significant improvement in accuracy over binary classification.


\section{The Proposed MT-DNN Model}
\label{sec:mt-dnn}
\begin{figure*}
	\centering
	\vspace{-1mm}
    {
	\includegraphics[width=0.92\textwidth]{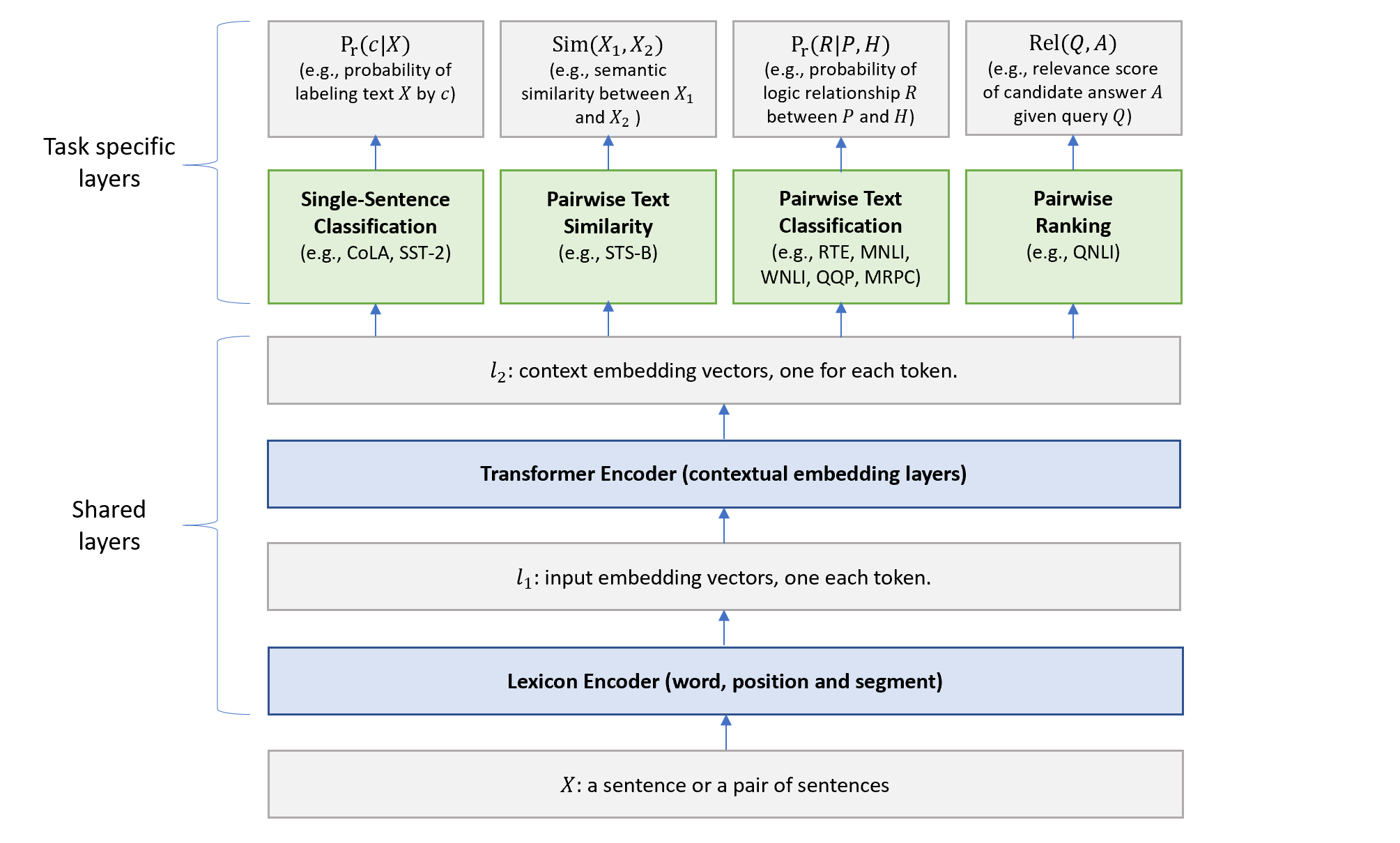}
    }
	\caption{Architecture of the MT-DNN model for representation learning. The lower layers are shared across all tasks while the top layers are task-specific. The input $X$ (either a sentence or a pair of sentences) is first represented as a sequence of embedding vectors, one for each word, in $l_1$. Then the Transformer encoder captures the contextual information for each word and generates the shared contextual embedding vectors in $l_2$. Finally, for each task, additional task-specific layers generate task-specific representations, followed by operations necessary for classification, similarity scoring, or relevance ranking.}
	\label{fig:mt-dnn}
\end{figure*}

The architecture of the MT-DNN model is shown in Figure \ref{fig:mt-dnn}. The lower layers are shared across all tasks, while the top layers represent task-specific outputs. The input $X$, which is a word sequence (either a sentence or a pair of sentences packed together) is first represented as a sequence of embedding vectors, one for each word, in $l_1$. Then the transformer encoder captures the contextual information for each word via self-attention, and generates a sequence of contextual embeddings in $l_2$. This is the shared semantic representation that is trained by our multi-task objectives.  In what follows, we elaborate on the model in detail.

\paragraph{Lexicon Encoder ($l_1$):} 
The input $X=\{x_1,...,x_m\}$ is a sequence of tokens of length $m$. Following \citet{bert2018}, the first token $x_1$ is always the \texttt{[CLS]} token. 
If $X$ is packed by a sentence pair $(X_1, X_2)$, we separate the two sentences with a special token \texttt{[SEP]}. The lexicon encoder maps $X$ into a sequence of input embedding vectors, one for each token, constructed by summing the corresponding word, segment, and positional embeddings.

\paragraph{Transformer Encoder ($l_2$):}
We use a multi-layer bidirectional Transformer encoder \citep{vaswani2017attention} to map the input representation vectors ($l_1$) into a sequence of contextual embedding vectors 
$\mathbf{C} \in \mathbb{R}^{d \times m}$. 
This is the shared representation across different tasks. Unlike the BERT model \citep{bert2018} that learns the representation via pre-training,
MT-DNN learns the representation using multi-task objectives, in addition to pre-training.

Below, we will describe the task specific layers using the NLU tasks in GLUE as examples, although in practice we can incorporate arbitrary natural language tasks such as text generation where the output layers are implemented as a neural decoder.

\paragraph{Single-Sentence Classification Output:}
Suppose that $\mathbf{x}$ is the contextual embedding ($l_2$) of the token \texttt{[CLS]}, which can be viewed as the semantic representation of input sentence $X$. Take the SST-2 task as an example. The probability that $X$ is labeled as class $c$ (i.e., the sentiment) is predicted by a logistic regression with softmax:
\begin{equation}
P_r(c|X)= \text{softmax} (\mathbf{W}_{SST}^\top \cdot \mathbf{x}),
\label{eqn:single-sent-classification}
\end{equation}
where $\mathbf{W}_{SST}$ is the task-specific parameter matrix.

\paragraph{Text Similarity Output:}
Take the STS-B task as an example. Suppose that $\mathbf{x}$ is the contextual embedding ($l_2$) of \texttt{[CLS]} which can be viewed as the semantic representation of the input sentence pair $(X_1, X_2)$. We introduce a task-specific parameter vector $\mathbf{w}_{STS}$ to compute the similarity score as: 
\begin{equation}
\text{Sim} (X_1, X_2)= \mathbf{w}_{STS}^\top \cdot \mathbf{x},
\label{eqn:text-sim}
\end{equation}
where $\text{Sim} (X_1, X_2)$ is a real value of the range (-$\infty$, $\infty$).


\paragraph{Pairwise Text Classification Output:}
Take natural language inference (NLI) as an example. The NLI task defined here involves a premise $P = (p_1,...,p_m)$ of $m$ words and a hypothesis $H = (h_1,..., h_n)$  of $n$ words, and aims to find a logical relationship $R$ between $P$ and $H$. The design of the output module follows the answer module of the stochastic answer network (SAN) \citep{liu2018san4nli}, a state-of-the-art neural NLI model. SAN's answer module uses multi-step reasoning. Rather than directly predicting the entailment given the input, it maintains a state and iteratively refines its predictions.

The SAN answer module works as follows. We first construct the working memory of premise $P$ by concatenating the contextual embeddings of the words in $P$, which are the output of the transformer encoder, denoted as $\mathbf{M}^p \in \mathbb{R}^{d \times m}$, and similarly the working memory of hypothesis $H$, denoted as $\mathbf{M}^h \in \mathbb{R}^{d \times n}$. 
Then, we perform $K$-step reasoning on the memory to output the relation label, where $K$ is a hyperparameter.
At the beginning,  the initial state $\mathbf{s}^0$ is the summary of $\mathbf{M}^h$: 
$\mathbf{s}^0 = \sum_j \alpha_j\mathbf{M}_j^h$, 
where $\alpha_j = \frac {\exp(\mathbf{w}_1^\top \cdot \mathbf{M}_j^h)} {\sum_i \exp(\mathbf{w}_1^\top \cdot \mathbf{M}_i^h)}$. 
At time step $k$ in the range of $\{1,2,…,K-1\}$, the state is defined by 
$\mathbf{s}^k = \text{GRU} (\mathbf{s}^{k-1}, \mathbf{x}^k)$. 
Here, $\mathbf{x}^k$ is computed from the previous state $\mathbf{s}^{k-1}$ and memory $\mathbf{M}^p$: $\mathbf{x}^k = \sum_j \beta_j\mathbf{M}_j^p$ and $\beta_j = \text{softmax} (\mathbf{s}^{k-1} \mathbf{W}_2^\top \mathbf{M}^p )$. 
A one-layer classifier is used to determine the relation at each step $k$:
\begin{equation}
P_r^k = \text{softmax} (\mathbf{W}_3^\top [\mathbf{s}^k ; \mathbf{x}^k ; |\mathbf{s}^k - \mathbf{x}^k|; \mathbf{s}^k \cdot \mathbf{x}^k ]).
\label{eqn:pairwise-text-classification}
\end{equation}
 
At last, we utilize all of the $K$ outputs by averaging the scores:
\begin{equation}
P_r = \text{avg} ([P_r^0, P_r^1, ..., P_r^{K-1}]).
\label{eqn:pairwise-text-classification-avg}
\end{equation}

Each $P_r$ is a probability distribution over all the relations $R \in \mathcal{R}$. 
During training, we apply \emph{stochastic prediction dropout} \citep{liu2018san} before the above averaging operation. 
During decoding, we average all outputs to improve robustness.

\paragraph{Relevance Ranking Output:}
Take QNLI as an example. Suppose that $\mathbf{x}$ is the contextual embedding vector of \texttt{[CLS]} which is the semantic representation of a pair of question and its candidate answer $(Q, A)$. 
We compute the relevance score as: 
\begin{equation}
\text{Rel} (Q, A)= g (\mathbf{w}_{QNLI}^\top \cdot \mathbf{x}),
\label{eqn:rel-score}
\end{equation}
For a given $Q$, we rank all of its candidate answers based on their relevance scores computed using Equation \ref{eqn:rel-score}.

\subsection{The Training Procedure}

The training procedure of MT-DNN consists of two stages: pretraining and multi-task learning. The pretraining stage follows that of the BERT model \citep{bert2018}. The parameters of the lexicon encoder and Transformer encoder are learned using two unsupervised prediction tasks: masked language modeling and next sentence prediction.\footnote{In this study we use the pre-trained BERT models released by the authors.}

In the multi-task learning stage, we use mini-batch based stochastic gradient descent (SGD) to learn the parameters of our model (i.e., the parameters of all shared layers and task-specific layers) as shown in Algorithm \ref{algo:mtdnn}.  In each epoch, a mini-batch $b_t$ is selected(e.g., among all 9 GLUE tasks), and the model is updated according to the task-specific objective for the task $t$. This approximately optimizes the sum of all multi-task objectives. 
\begin{algorithm}[ht!]
 \SetAlgoLined
Initialize model parameters $\Theta$ randomly.  \\
Pre-train the shared layers (i.e., the lexicon encoder and the transformer encoder). \\
Set the max number of epoch: $epoch_{max}$.
\textit{//Prepare the data for $T$ tasks.}\\
\For{$t$ in $1,2,...,T$ }
{
    Pack the dataset $t$ into mini-batch: $D_t$.
}

 \For{$epoch$ in $1,2,...,epoch_{max}$}{
     1. Merge all the datasets: $D =D_1 \cup D_2 ... \cup D_T$ \\
     2. Shuffle $D$ \\
     \For{$b_t$ in D}{
        \textit{//$b_t$ is a mini-batch of task $t$.} \\
     3. Compute loss : $L(\Theta)$ \\
        \hspace{0.3cm} $L(\Theta)=$ Eq.~\ref{eqn:cross-entropy-loss} for classification \\
        \hspace{0.3cm} $L(\Theta)=$ Eq.~\ref{eqn:msq-loss} for regression \\
        \hspace{0.3cm} $L(\Theta)=$ Eq.~\ref{eqn:ranking-loss} for ranking \\
     4. Compute gradient: $\nabla(\Theta)$ \\
     5. Update model: $\Theta = \Theta - \epsilon \nabla(\Theta)$ \\
     }
 }
 \caption{\label{algo:mtdnn} Training a MT-DNN model.}
\end{algorithm} 



For the classification tasks (i.e., single-sentence or pairwise text classification), we use the cross-entropy loss as the objective:
\begin{equation}
-\sum_c \mathbbm{1}(X,c) \log(P_r(c|X)),
\label{eqn:cross-entropy-loss}
\end{equation}
where $\mathbbm{1}(X,c)$ is the binary indicator (0 or 1) if class label $c$ is the correct classification for $X$, and $P_r(.)$ is defined by e.g., Equation \ref{eqn:single-sent-classification} or \ref{eqn:pairwise-text-classification-avg}.

For the text similarity tasks, such as STS-B, where each sentence pair is annotated with a real-valued score $y$, we use the mean squared error as the objective:
\begin{equation}
(y - \text{Sim}(X_1, X_2))^2,
\label{eqn:msq-loss}
\end{equation}
where $\text{Sim}(.)$ is defined by Equation \ref{eqn:text-sim}.

The objective for the relevance ranking tasks follows the pairwise learning-to-rank paradigm \citep{learning-to-rank2005burges,huang2013dssm}. Take QNLI as an example. Given a query $Q$, we obtain a list of candidate answers $\mathcal{A}$ which contains a positive example $A^+$ that includes the correct answer, and $|\mathcal{A}|-1$ negative examples. We then minimize the negative log likelihood of the positive example given queries across the training data
\begin{equation}
-\sum_{(Q,A^+)} P_r(A^+ | Q),
\label{eqn:ranking-loss}
\end{equation}
\begin{equation}
P_r(A^+ | Q) = \frac{\exp(\gamma \text{Rel}(Q,A^+))}{\sum_{A^{'} \in \mathcal{A}} \exp(\gamma \text{Rel}(Q,A^{'}))},
\label{eqn:ranking-prob}
\end{equation}
where $\text{Rel}(.)$ is defined by Equation \ref{eqn:rel-score} and $\gamma$ is a tuning factor determined on held-out data. In our experiment, we simply set $\gamma$ to 1.


\section{Experiments}
\label{sec:exp}
\begin{table*}[htb!]
	\begin{center}
		\begin{tabular}{l|l|c|c|c|c|c}
			\hline \bf Corpus &Task& \#Train & \#Dev & \#Test   & \#Label &Metrics\\ \hline \hline
			\multicolumn{6}{@{\hskip1pt}r@{\hskip1pt}}{Single-Sentence Classification (GLUE)} \\ \hline
			CoLA & Acceptability&8.5k & 1k & 1k & 2 & Matthews corr\\ \hline
			SST-2 & Sentiment&67k & 872 & 1.8k & 2 & Accuracy\\ \hline \hline
			\multicolumn{6}{@{\hskip1pt}r@{\hskip1pt}}{Pairwise Text Classification (GLUE)} \\ \hline
			MNLI & NLI& 393k& 20k & 20k& 3 & Accuracy\\ \hline
            RTE & NLI &2.5k & 276 & 3k & 2 & Accuracy \\ \hline
            WNLI & NLI &634& 71& 146& 2 & Accuracy \\ \hline
			QQP & Paraphrase&364k & 40k & 391k& 2 & Accuracy/F1\\ \hline
            MRPC & Paraphrase &3.7k & 408 & 1.7k& 2&Accuracy/F1\\ \hline
			\multicolumn{5}{@{\hskip1pt}r@{\hskip1pt}}{Text Similarity (GLUE)} \\ \hline
			STS-B & Similarity &7k &1.5k& 1.4k &1 & Pearson/Spearman corr\\ \hline

\multicolumn{6}{@{\hskip1pt}r@{\hskip1pt}}{Relevance Ranking (GLUE)} \\ \hline \hline
			QNLI & QA/NLI& 108k &5.7k&5.7k&2& Accuracy\\ \hline \hline
			\multicolumn{6}{@{\hskip1pt}r@{\hskip1pt}}{Pairwise Text Classification} \\ \hline
			SNLI & NLI& 549k &9.8k&9.8k&3& Accuracy\\ \hline
			SciTail & NLI& 23.5k &1.3k&2.1k&2& Accuracy\\ \hline

		\end{tabular}
	\end{center}
	\caption{Summary of the three benchmarks: GLUE, SNLI and SciTail.
	}
	\label{tab:datasets}
\end{table*}

We evaluate the proposed {\MNAME} on three popular NLU benchmarks: GLUE \cite{wang2018glue}, SNLI \cite{snli2015}, and SciTail \cite{scitail}. 
We compare MT-DNN with existing state-of-the-art models including BERT 
and demonstrate the effectiveness of MTL with and without model fine-tuning using GLUE and domain adaptation using both SNLI and SciTail. 

\subsection{Datasets}
\label{subsec:dataset}
This section briefly describes the GLUE, SNLI, and SciTail datasets, as summarized in Table~\ref{tab:datasets}.

\paragraph{GLUE} The General Language Understanding Evaluation (GLUE) benchmark is a collection of nine NLU tasks as in Table~\ref{tab:datasets}, including question answering, sentiment analysis, text similarity and textual entailment; it is considered well-designed for evaluating the generalization and robustness of NLU models. 

\paragraph{SNLI}
The Stanford Natural Language Inference (SNLI) dataset contains 570k human annotated sentence pairs, in which the premises are drawn from the captions of the Flickr30 corpus and hypotheses are manually annotated \cite{snli2015}. 
This is the most widely used entailment dataset for NLI.
The dataset is used only for domain adaptation in this study.

\paragraph{SciTail}
This is a textual entailment dataset derived from a science question answering (SciQ) dataset \cite{scitail}. The task involves assessing whether a given premise entails a given hypothesis.  
In contrast to other entailment datasets mentioned previously, the hypotheses in SciTail are created from science questions while the corresponding answer candidates and premises come from relevant web sentences retrieved from a large corpus. As a result, these sentences are linguistically challenging and the lexical similarity of premise and hypothesis is often high, thus making SciTail particularly difficult. 
The dataset is used only for domain adaptation in this study.

\subsection{Implementation details}
\label{subsec:impl}
Our implementation of MT-DNN is based on the PyTorch implementation of BERT\footnote{https://github.com/huggingface/pytorch-pretrained-BERT}.
We used Adamax \cite{kingma2014adam} as our optimizer with a learning rate of 5e-5 and a batch size of 32 by following \citet{bert2018}. 
The maximum number of epochs was set to 5. 
A linear learning rate decay schedule with warm-up over 0.1 was used, unless stated otherwise. 
We also set the dropout rate of all the task specific layers as 0.1, except 0.3 for MNLI and 0.05 for CoLa. 
To avoid the exploding gradient problem, we clipped the gradient norm within 1. 
All the texts were tokenized using wordpieces, and were chopped to spans no longer than 512 tokens.


\begin{table*}[htb!]
\small
	\begin{center}
		\begin{tabular}{l|@{\hskip1pt}l@{\hskip1pt}|@{\hskip1pt}c@{\hskip1pt}|@{\hskip1pt}c@{\hskip1pt}|@{\hskip1pt}c@{\hskip1pt}|@{\hskip1pt}c@{\hskip1pt}|@{\hskip1pt}c|@{\hskip1pt}c|@{\hskip1pt}c |@{\hskip1pt} c |@{\hskip1pt} c|@{\hskip1pt} c}
			\hline \bf Model &CoLA&	SST-2 &MRPC& STS-B&QQP&MNLI-m/mm&QNLI&RTE&WNLI&AX &\textbf{Score}\\ 
			& 8.5k &67k &3.7k &7k &364k &393k &108k &2.5k &634 & & \\ \hline \hline
			BiLSTM+ELMo+Attn $^1$&36.0 &90.4 &84.9/77.9 &75.1/73.3 &64.8/84.7 &76.4/76.1 &- &56.8 &65.1 &26.5 &70.5 \\ \hline
			\begin{tabular}{@{}c@{}}Singletask Pretrain \\Transformer $^2$   \end{tabular}
			 &45.4 &91.3 &82.3/75.7&82.0/80.0 &70.3/88.5 &82.1/81.4 &- &56.0 &53.4  &29.8 &72.8 \\ \hline
			GPT on STILTs $^3$ &47.2 &93.1 &87.7/83.7 &85.3/84.8 &70.1/88.1 &80.8/80.6 &- &69.1 &65.1 &29.4 &76.9 \\ \hline
			BERT$_{\text{LARGE}}^4$ & 60.5 &94.9 &89.3/85.4 &87.6/86.5 &72.1/89.3 &86.7/85.9 &92.7 &70.1 &65.1	&39.6 & 80.5\\ \hline
            {\MNAME}\textsubscript{no-fine-tune} & 58.9 &94.6 &\color{blue}{\textbf{90.1/86.4}} &89.5/88.8 &\color{blue}{\textbf{72.7/89.6}} &86.5/85.8	&\color{blue}{\textbf{93.1}} &79.1 &65.1	&39.4 &81.7 \\ \hline \hline
			{\MNAME} & \textbf{62.5} &\textbf{95.6} &\color{blue}{\textbf{91.1/88.2}} &\textbf{89.5/88.8} &\color{blue}{\textbf{72.7/89.6}} &\textbf{86.7/86.0}	&\color{blue}{\textbf{93.1}} &\textbf{81.4} &65.1	&\textbf{40.3} &\textbf{82.7} \\ \hline \hline
			{Human Performance} &66.4&97.8&86.3/80.8    &92.7/92.6	&59.5/80.4	&92.0/92.8	&91.2	&93.6	&95.9 &- & 87.1\\ \hline
		\end{tabular}
	\end{center}
	\caption{GLUE test set results scored using the GLUE evaluation server. The number below each task denotes the number of training examples. The state-of-the-art results are in \textbf{bold}, and the results on par with or pass human performance are in {\color{blue}\textbf{bold}}. MT-DNN uses BERT\textsubscript{LARGE} to initialize its shared layers. 
	All the results are obtained from \href{https://gluebenchmark.com/leaderboard}{https://gluebenchmark.com/leaderboard} on February 25, 2019. 
	Model references: $^1$:\protect\cite{wang2018glue} ; $^2$:\protect\cite{gpt2018}; $^3$: \protect\cite{phang2018sentence}; $^4$:\protect\cite{bert2018}.
	}
	\label{tab:glue_test}
\end{table*}
\begin{table*}[h!]
	\begin{center}
		\begin{tabular}{l|c@{\hskip1pt}|c@{\hskip1pt}|c@{\hskip1pt}|c@{\hskip1pt}|c@{\hskip1pt}|@{\hskip1pt}c @{\hskip1pt}|c @{\hskip1pt}|c@{\hskip1pt}}
			\hline \bf Model &MNLI-{m/mm} & QQP & RTE & QNLI (v1/v2)  &MRPC & CoLa &SST-2  & STS-B \\ \hline \hline
			BERT$_{\text{LARGE}}$& 86.3/86.2 &91.1/88.0 &71.1 &90.5/92.4 &89.5/85.8 &61.8 &93.5 &89.6/89.3\\
			\hline
			ST-DNN &86.6/86.3 & 91.3/88.4 &  72.0& 96.1/- & 89.7/86.4 &- &- &-\\ \hline
            {\MNAME}  &\textbf{87.1/86.7} &\textbf{91.9/89.2} &\textbf{83.4}&\textbf{97.4/92.9} &\textbf{91.0/87.5} &\textbf{63.5}& \textbf{94.3}&\textbf{90.7/90.6} \\ \hline

		\end{tabular}
	\end{center}
	\caption{GLUE dev set results. The best result on each task is in \textbf{bold}. 
	The Single-Task DNN (ST-DNN) uses the same model architecture as MT-DNN. But its shared layers are the pre-trainedBERT model without being refined via MTL. We fine-tuned ST-DNN for each GLUE task using task-specific data.
	There have been two versions of the QNLI dataset. V1 is expired on January 30, 2019. The current version is v2.
	MT-DNN use BERT\textsubscript{LARGE} as their initial shared layers. 
	}
	\label{tab:glue_dev}
\end{table*}

\subsection{GLUE Main Results}
\label{subsec:results}
We compare MT-DNN with its variants and a list of state-of-the-art models that have been submitted to the GLUE leaderboard. The results are shown in Tables \ref{tab:glue_test} and \ref{tab:glue_dev}. 

\paragraph{BERT\textsubscript{LARGE}} This is the large BERT model released by the authors, which we used as a baseline. We fine-tuned the model for each GLUE task on task-specific data.

\paragraph{MT-DNN} This is the proposed model described in Section 3. We used the pre-trained BERT\textsubscript{LARGE} to initialize its shared layers, refined the model via MTL on all GLUE tasks, and fine-tuned the model for each GLUE task using task-specific data. The test results in Table~\ref{tab:glue_test} show that MT-DNN outperforms all existing systems on all tasks, except WNLI, creating new state-of-the-art results on eight GLUE tasks and pushing the benchmark to 82.7\%, which amounts to 2.2\% absolution improvement over BERT\textsubscript{LARGE}. Since MT-DNN uses BERT\textsubscript{LARGE} to initialize its shared layers, the gain is mainly attributed to the use of MTL in refining the shared layers. 
MTL is particularly useful for the tasks with little in-domain training data. 
As we observe in the table, on the same type of tasks, the improvements over BERT are much more substantial for the tasks with less in-domain training data than those with more in-domain labels, even though they belong to the same task type, e.g., the two NLI tasks: RTE vs. MNLI, and the two paraphrase tasks: MRPC vs. QQP.

\paragraph{MT-DNN\textsubscript{no-fine-tune}} Since the MTL of MT-DNN uses all GLUE tasks, it is possible to directly apply MT-DNN to each GLUE task without fine-tuning.
The results in Table 2 show that MT-DNN\textsubscript{no-fine-tune} still outperforms BERT\textsubscript{LARGE} consistently among all tasks but CoLA. Our analysis shows that CoLA is a challenge task with much smaller in-domain data than other tasks, and its task definition and dataset are unique among all GLUE tasks, making it difficult to benefit from the knowledge learned from other tasks. As a result, MTL tends to underfit the CoLA dataset. In such a case, fine-tuning is necessary to boost the performance. As shown in Table~\ref{tab:glue_test}, the accuracy improves from 58.9\% to 62.5\% after fine-tuning, even though only a very small amount of in-domain data is available for adaptation. This, together with the fact that the fine-tuned MT-DNN significantly outperforms the fine-tuned BERT\textsubscript{LARGE} on CoLA (62.5\% vs. 60.5\%), reveals that the learned MT-DNN representation allows much more effective domain adaptation than the pre-trained BERT representation. We will revisit this topic with more experiments in Section~\ref{subsec:domain}.

The gain of MT-DNN is also attributed to its flexible modeling framework which allows us to incorporate the task-specific model structures and training methods which have been developed in the single-task setting, effectively leveraging the existing body of research. Two such examples are the use of the SAN answer module for the pairwise text classification output module and the pairwise ranking loss for the QNLI task which by design is a binary classification problem in GLUE. To investigate the relative contributions of these modeling design choices, we implement a variant of MT-DNN as described below.

\paragraph{ST-DNN} ST-DNN stands for Single-Task DNN. It uses the same model architecture as MT-DNN. But its shared layers are the pre-trained BERT model without being refined via MTL. We then fine-tuned ST-DNN for each GLUE task using task-specific data. Thus, for pairwise text classification tasks, the only difference between their ST-DNNs and BERT models is the design of the task-specific output module. The results in Table~\ref{tab:glue_dev} show that on all four tasks (MNLI, QQP, RTE and MRPC) ST-DNN outperforms BERT, justifying the effectiveness of the SAN answer module. We also compare the results of ST-DNN and BERT on QNLI. While ST-DNN is fine-tuned using the pairwise ranking loss, BERT views QNLI as binary classification and is fine-tuned using the cross entropy loss. ST-DNN significantly outperforms BERT demonstrates clearly the importance of problem formulation.

\subsection{Domain Adaptation Results on SNLI and SciTail}
\label{subsec:domain}

\begin{figure}[h!]
    \centering
 {
	\includegraphics[width=0.48\textwidth]{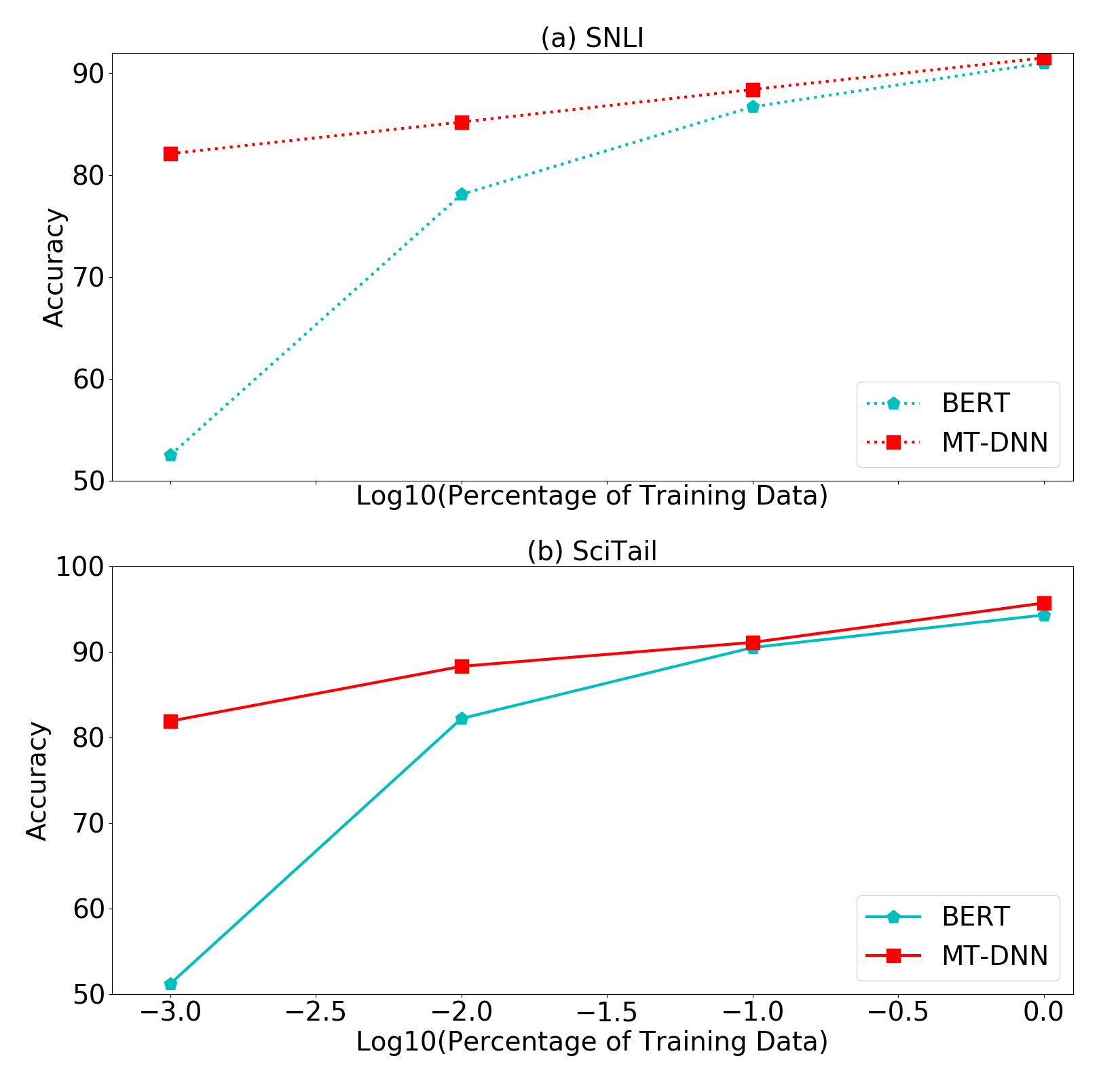}
    }   
    \caption{\label{fig:domain} Domain adaption results on SNLI and SciTail development datasets using the shared embeddings generated by MT-DNN and BERT, respectively. Both MT-DNN and BERT are fine-tuned based on the pre-trained BERT$_\text{BASE}$. The X-axis indicates the amount of domain-specific labeled samples used for adaptation.}
\end{figure}
\begin{table}[htb!]
	\begin{center}
		\begin{tabular}{@{\hskip1pt}l@{\hskip1pt} |@{\hskip1pt} c |@{\hskip1pt} c |@{\hskip1pt} c | c@{\hskip1pt}}
			\hline \bf Model & 0.1\% & 1\% &10\% & 100\% \\ \hline
            \multicolumn{5}{c}{ SNLI Dataset (Dev Accuracy\%)} \\ \hline
            \#Training Data &549& 5,493& 54,936&549,367 \\ \hline
            BERT &52.5&78.1&86.7 & 91.0 \\ \hline
            MT-DNN &82.1 & 85.2 & 88.4 & 91.5 \\ \hline \hline
\multicolumn{5}{c}{ SciTail Dataset (Dev Accuracy\%)} \\ \hline
            \#Training Data &23& 235& 2,359& 23,596\\ \hline
            BERT &51.2&82.2&90.5 & 94.3 \\ \hline
            MT-DNN &81.9 & 88.3 & 91.1 & 95.7 \\ \hline			
		\end{tabular}
	\end{center}
	\caption{Domain adaptation results on SNLI and SciTail, as shown in Figure~\ref{fig:domain}.  
	}
	\label{tab:domain}
\end{table}

One of the most important criteria of building practical systems is fast adaptation to new tasks and domains. This is because it is prohibitively expensive to collect labeled training data for new domains or tasks. Very often, we only have very small training data or even no training data.

To evaluate the models using the above criterion, we perform domain adaptation experiments on two NLI tasks, SNLI and SciTail, using the following procedure:
\begin{enumerate}
    \item use the MT-DNN model or the BERT as initial model including both \textit{BASE} and \textit{LARGE} model settings;
    \item create for each new task (SNLI or SciTail) a task-specific model, by adapting the trained MT-DNN using task-specific training data;
    \item evaluate the models using task-specific test data.
\end{enumerate}

We starts with the default training/dev/test set of these tasks. But we randomly sample 0.1\%, 1\%, 10\% and 100\% of its training data. As a result, we obtain four sets of training data for SciTail, which respectively includes  23, 235, 2.3k and 23.5k training samples. Similarly, we obtain four sets of training data for SNLI, which respectively include 549, 5.5k, 54.9k and 549.3k training samples.

We perform  random sampling five times and report the mean among all the runs. Results on different amounts of training data from SNLI and SciTail are reported in Figure~\ref{fig:domain}. We observe that MT-DNN  outperforms the BERT baseline consistently with more details provided in Table \ref{tab:domain}. 
The fewer training examples used, the larger improvement MT-DNN demonstrates over BERT. For example, with only 0.1\% (23 samples) of the SNLI training data, MT-DNN achieves 82.1\% in accuracy while BERT's accuracy is 52.5\%; with 1\%\ of the training data, the accuracy from MT-DNN is 85.2\% and BERT is 78.1\%. We observe similar results on SciTail. 
The results indicate that the representations learned by MT-DNN are more consistently effective for domain adaptation than BERT.

In Table~\ref{tab:nli}, we compare our adapted models, using all in-domain training samples, against several strong baselines including the best results reported in the leaderboards. We see that MT-DNN\textsubscript{LARGE} generates new state-of-the-art results on both datasets, pushing the benchmarks to 91.6\% on SNLI (1.5\% absolute improvement) and 95.0\% on SciTail (6.7\% absolute improvement), respectively. This results in the new state-of-the-art for both SNLI and SciTail. All of these demonstrate the exceptional performance of MT-DNN on domain adaptation.
\begin{table}[htb!]
	\begin{center}
		\begin{tabular}{l | c | c }\hline
	   \bf Model &Dev& Test  \\ \hline 

		\multicolumn{3}{c}{ SNLI Dataset (Accuracy\%)}  \\ \hline 
		GPT \cite{gpt2018} &- & 89.9 \\
		\hline
		\citet{kim2018semantic}$^*$ &- &90.1 \\ \hline 
		BERT\textsubscript{BASE} &91.0 & 90.8 \\
		\hline		
		MT-DNN\textsubscript{BASE} &91.5 & 91.1 \\
		\hline
		BERT\textsubscript{LARGE} &91.7& 91.0\\ \hline		
		{\MNAME}\textsubscript{LARGE}&\textbf{92.2}& \textbf{91.6}\\ \hline		
		\hline
		\multicolumn{3}{c}{ SciTail Dataset (Accuracy\%)}  \\ \hline 	GPT \cite{gpt2018}$^*$ &- &88.3 \\ \hline
		BERT\textsubscript{BASE} &94.3 & 92.0 \\ \hline
		MT-DNN\textsubscript{BASE} &95.7 &94.1 \\
		\hline
		BERT\textsubscript{LARGE} &95.7& 94.4\\ \hline		
		{\MNAME}\textsubscript{LARGE} &\textbf{96.3}& \textbf{95.0}\\ \hline		
		\end{tabular}
	\end{center}
	\caption{Results on the SNLI and SciTail dataset. Previous state-of-the-art results are marked by $*$, obtained from the official SNLI leaderboard (https://nlp.stanford.edu/projects/snli/) and the official SciTail leaderboard maintained by AI2 (https://leaderboard.allenai.org/scitail). 
	}
	\label{tab:nli}

\end{table}

\section{Conclusion}
\label{sec:con}
In this work we proposed a model called MT-DNN to combine multi-task learning and language model pre-training for language representation learning. 
MT-DNN obtains new state-of-the-art results on ten NLU tasks across three popular benchmarks: SNLI, SciTail, and GLUE.
MT-DNN also demonstrates an exceptional generalization capability in domain adaptation experiments. 

There are many future areas to explore to improve MT-DNN, including a deeper understanding of model structure sharing in MTL, a more effective training method that leverages relatedness among multiple tasks, for both fine-tuning and pre-training \cite{unilm2019}, and ways of incorporating the linguistic structure of text in a more explicit and controllable manner. At last, we also would like to verify whether MT-DNN is resilience against adversarial attacks \cite{breaknli2019acl,talman2018testing,liu2019mt-dnn-kd}.


\section*{Acknowledgments}
\label{sec:akn}

We would like to thanks Jade Huang from Microsoft for her generous help on this work.


\bibliography{acl_snli}
\bibliographystyle{acl_natbib}
%

\end{document}


\onecolumn

\section{Appendices}
\label{sec:appendix}


\subsection{Test results on the \textit{old} GLUE test set}
\label{sec:glue_old}
\begin{table*}[h!]
\small
	\begin{center}
		\begin{tabular}{l|@{\hskip1pt}l@{\hskip1pt}|@{\hskip1pt}c@{\hskip1pt}|@{\hskip1pt}c@{\hskip1pt}|@{\hskip1pt}c@{\hskip1pt}|@{\hskip1pt}c@{\hskip1pt}|@{\hskip1pt}c|@{\hskip1pt}c|@{\hskip1pt}c |@{\hskip1pt} c |@{\hskip1pt} c|@{\hskip1pt} c}
			\hline \bf Model &CoLA&	SST-2 &MRPC& STS-B&QQP&MNLI-m/mm&QNLI&RTE&WNLI&AX &\textbf{Score}\\ 
			& 8.5k &67k &3.7k &7k &364k &393k &108k &2.5k &634 & & \\ \hline \hline
			BiLSTM+ELMo+Attn $^1$ &36.0 &90.4 &84.9/77.9 &75.1/73.3 &64.8/84.7 &76.4/76.1 &79.9 &56.8 &65.1 &26.5 &70.5 \\ \hline
			\begin{tabular}{@{}c@{}}Singletask Pretrain \\Transformer $^2$  \end{tabular}
			 &45.4 &91.3 &82.3/75.7&82.0/80.0 &70.3/88.5 &82.1/81.4 &88.1 &56.0 &53.4  &29.8 &72.8 \\ \hline
			GPT on STILTs $^3$ &47.2 &93.1 &87.7/83.7 &85.3/84.8 &70.1/88.1 &80.8/80.6 &87.2 &69.1 &65.1 &29.4 &76.9 \\ \hline
			BERT$_{\text{LARGE}}$ $^4$ & 60.5 &94.9 &89.3/85.4 &87.6/86.5 &72.1/89.3 &86.7/85.9 &91.1 &70.1 &65.1	&39.6 & 80.4\\ \hline
MT-DNN &\textbf{61.5} &\textbf{95.6} &\textbf{90.0/86.7} &\textbf{88.3/87.7} &\textbf{72.4/89.6} &\textbf{86.7/86.0}	&\textbf{98.0} &\textbf{75.5} &65.1	&\textbf{40.3} &\textbf{82.2} \\ \hline
		\end{tabular}
	\end{center}
	\caption{GLUE test set results, which are scored by the GLUE evaluation server. The numbers below each task denote the size of training examples. The state-of-the-art results are in \textbf{bold}. All the results are obtained from \href{https://gluebenchmark.com/leaderboard}{https://gluebenchmark.com/leaderboard} on January 15, 2019. Note that the \textit{old} version of GLUE test set expired on January 30, 2019. Model references: $^1$:\protect\cite{wang2018glue} ; $^2$:\protect\cite{gpt2018}; $^3$: \protect\cite{phang2018sentence};  $^4$:\protect\cite{bert2018}.
	}
	\label{tab:glue_test}
\end{table*}
\bibliography{acl_snli}
\bibliographystyle{acl_natbib}